\documentclass[journal]{IEEEtran}
\usepackage{graphicx} 
\usepackage{caption}
\usepackage{cite}
\usepackage[numbers,sort&compress]{natbib}
\usepackage{amssymb}
\usepackage{subcaption}
\usepackage{multirow}
\usepackage{listings}
\usepackage{xcolor}
\usepackage{booktabs}
\usepackage[framemethod=TikZ]{mdframed}
\usepackage{svg}
\usepackage{enumerate}

\usepackage{amsmath}
\DeclareMathOperator*{\argmax}{argmax}
\newcommand\vect[1]{\boldsymbol{#1}}
\usepackage{float}
\usepackage{tabularx}
\usepackage{algorithm}
\usepackage{algpseudocode}
\usepackage{hyperref}
\usepackage{makecell}

\usepackage{mathtools}
\DeclarePairedDelimiter\norm{\lVert}{\rVert}%

\begin{document}
\title{Optimizing Plastic Waste Collection in Water Bodies Using Heterogeneous Autonomous Surface Vehicles with Deep Reinforcement Learning}

\author{Alejandro Mendoza Barrionuevo$^{1}$, Samuel Yanes Luis$^{1}$, Daniel Gutiérrez Reina$^{1}$ and Sergio L. Toral Marín$^{1}$

\thanks{This work was funded by ``Junta de Andalucía: Consejería de Universidad, Investigación e Innovación'' through the project ``Monitorization of Environmental Dangers with Unmanned Surface Agents: (MEDUSA)'' under Grant PCM\_00019.}
\thanks{$^{1}$Department of Electronic Engineering, University of Sevilla, Av. de Los Descubrimientos s/n, Sevilla, 41003, Spain}
}

\markboth{IEEE Robotics and Automation Letters}{\hfill My Right Header}

\maketitle
\pagestyle{empty}  
\thispagestyle{empty} 

\begin{abstract}
This paper presents a model-free deep reinforcement learning framework for informative path planning with heterogeneous fleets of autonomous surface vehicles to locate and collect plastic waste. The system employs two teams of vehicles: scouts and cleaners. Coordination between these teams is achieved through a deep reinforcement approach, allowing agents to learn strategies to maximize cleaning efficiency.
The primary objective is for the scout team to provide an up-to-date contamination model, while the cleaner team collects as much waste as possible following this model. This strategy leads to heterogeneous teams that optimize fleet efficiency through inter-team cooperation supported by a tailored reward function.
Different trainings of the proposed algorithm are compared with other state-of-the-art heuristics in two distinct scenarios, one with high convexity and another with narrow corridors and challenging access. According to the obtained results, it is demonstrated that deep reinforcement learning based algorithms outperform other benchmark heuristics, exhibiting superior adaptability.
In addition, training with greedy actions further enhances performance, particularly in scenarios with intricate layouts.

\end{abstract}


\section{Introduction}
Every year, millions of tons of plastic end up in oceans, rivers and lakes, seriously affecting marine life and aquatic ecosystems. According to a recent report from the World Economic Forum \cite{plasticfish2016}, it is estimated that by 2050 there will be more plastic than fish by weight in the oceans if urgent action is not taken. Traditional cleaning methods, such as hand nets and manned boats, have usually been proven to be inefficient and poorly scalable.

This paper introduces an innovative approach that employs a fleet of ASVs designed specifically for the collective task of tracking and removing plastic waste from water bodies. The fleet consists of a heterogeneous group of ASVs, which is divided into two specialized teams: cleaning vehicles and scout vehicles. These need to cooperate to maximize cleaning efficiency, resulting in a heterogeneous multirobot system (HMRS). Scout vehicles, designed to be lighter and faster, are equipped with cameras capable of detecting and accurately mapping the location of plastic waste over a wide range. On the other hand, cleaner vehicles are slower due to the additional trash collection systems they carry, and are equipped with less advanced cameras to reduce costs. These cameras allow the detection of waste in a significantly smaller area compared to scout vehicles.

To coordinate cooperation between these vehicles and maximize plastic waste collection, this research focuses on the use of deep reinforcement learning (DRL). More specifically, a Deep Q-Learning algorithm is implemented in two deep neural networks (DNNs), each of them shared by the agents of the same team.
The main objective is that the scout team provides the most updated contamination model possible, especially in areas with high trash density, while the cleaner team is in charge of collecting as much trash as possible following this trash-density model. This strategy leads to the development of an informative path planning (IPP) system for heterogenous teams of ASVs, which can be defined as the process of generating paths for agents to efficiently gather information from an environment, optimizing certain factors. In this case, the policy must seek to optimize the cooperative efficiency of the cleaning task. Overall, the main contributions of this work are as follows:

\begin{enumerate}[i)]
    \item The development of a model-free DRL framework for IPP with heterogeneous fleets of ASVs to have intra-team and cross-team cooperation under realistic constraints.
    \item The definition of tailored observation and reward functions to drive the algorithm towards the collective goal. Through state representation and rewards, teams should learn to work cooperatively, since the better one does its task, the easier it will be for the other.
    \item To evaluate the performance of the design, a thorough analysis and comparison with other heuristics and algorithms in the literature is carried out.
\end{enumerate}

\section{Related Work}
\label{Related Work}

This work intersects various fields of study, namely multirobot systems, DRL and IPP for autonomous vehicles. The domain of multirobot systems has progressed significantly in recent times, driven by advances in computing and communication technologies \cite{advancesinmultirobotsurvey}. However, despite some research on heterogeneous robot systems \cite{rizk2019cooperative}, most multirobot studies focus on homogeneous configurations. 
Moreover, although DRL has been applied to multi-agent systems, its use is still more frequent in single-agent systems, as highlighted in surveys \cite{hernandez2019survey}.

Given the challenge of planning autonomous vehicle trajectories, several algorithms have been employed in IPP \cite{popovic2024learning}, such as bio-inspired algorithms \cite{micaela2022aquafelpso}, Bayesian Optimization \cite{Federico57.2023}, or DRL algorithms \cite{drlpathplanning, vashisth2024deep}. 

Among all of them, as seen in works like \cite{lei2018dynamic}, DRL has established itself as one of the most effective methods in the field of trajectory optimization based on information collected in real time, including multi-agent systems. In \cite{samuel2024deepGPS}, a homogeneous multi-agent system composed of ASVs is presented to efficiently monitor water quality, integrating local Gaussian processes with DRL techniques. 
In \cite{yang2020multi}, the authors propose a multirobot path planning model for warehouse dispatching system using an enhanced Deep Q-Network algorithm with prior knowledge and predefined conflict resolution rules to optimize robot interactions.

DRL has also been widely applied to other fields within robotics in recent years, in addition to path planning, such as collision avoidance \cite{woo2020collision}, or robotic systems control \cite{gu2017deep}, among others. Additionally, although to a lesser degree, related work on DRL applied to heterogeneous systems can be found. 
In \cite{barrionuevo2024informative}, Dueling Double Deep Q-Learning (DDDQL) is employed with prioritized experience replay to manage fleets of heterogeneous ASVs that carry measurement sensors of different qualities. The purpose is to obtain the best contamination model of the aquatic environment employing also Gaussian processes. 
DDDQL is also applied in \cite{calvo2018heterogeneoustrafficlights} to address heterogeneous multi-agent in urban traffic control environments, where each agent represents a city traffic junction. Reference \cite{gao2023asymmetric} focuses on creating an intelligent system for adversarial catching tasks, employing HMRS and multi-agent actor-critic DRL. To enhance cooperation among different types of robot, it introduces asymmetric self-play and curriculum learning techniques. 

Most studies addressing floating garbage cleaning robots primarily focus on the physical development of prototypes and the mechanical motion control of these systems, such as \cite{akib2019unmanned} or \cite{kamarudin2021development}. 
However, there is a notable gap in research regarding algorithms that guide agents at a high level for detection, positioning and garbage collection tasks. In \cite{deng2022automatic}, a heterogeneous collaborative system is proposed where aerial drones handle waste detection and ASVs perform collection tasks. Aerial drones use a partition-based coverage algorithm, while ASVs are tasked by particle swarm optimization (PSO) and guided by ant colony optimization for path planning. In contrast, our approach leverages DRL framework to autonomously manage decision-making of all type of agents without predefined task assignments, relying on ASVs for both cleaning and waste detection. Additionally, while their study assumes static waste, ours simulates dynamic waste movement, introducing environmental currents for added realism.

\section{Methodology}
\label{Methodology}

\subsection{Exploration and Cleaning}
This work addresses a waste collection problem in aquatic environments deploying a heterogeneous fleet of ASVs divided into two specialized teams with complementary roles: scouts and cleaners. The scout team consists of smaller, highly mobile and hydrodynamic ASVs equipped with wide-range vision systems, allowing them to cover large areas efficiently. However, these vehicles do not have the ability to collect trash. Their primary task is to identify and map the locations of floating waste, constantly updating a shared model, which is essential for the cleaning team. On the other hand, the cleaner team, comprises ASVs designed specifically to collect the trash. Due to their larger size and the weight of the collection equipment, these vehicles operate at a slower speed. For this reason, and to reduce costs, they are equipped with a simpler mono camera rather than the stereo cameras used by the scout team. This feature limits their detection range to only nearby trash.

\begin{figure}
    \centering
    \includegraphics[width=0.9\linewidth]{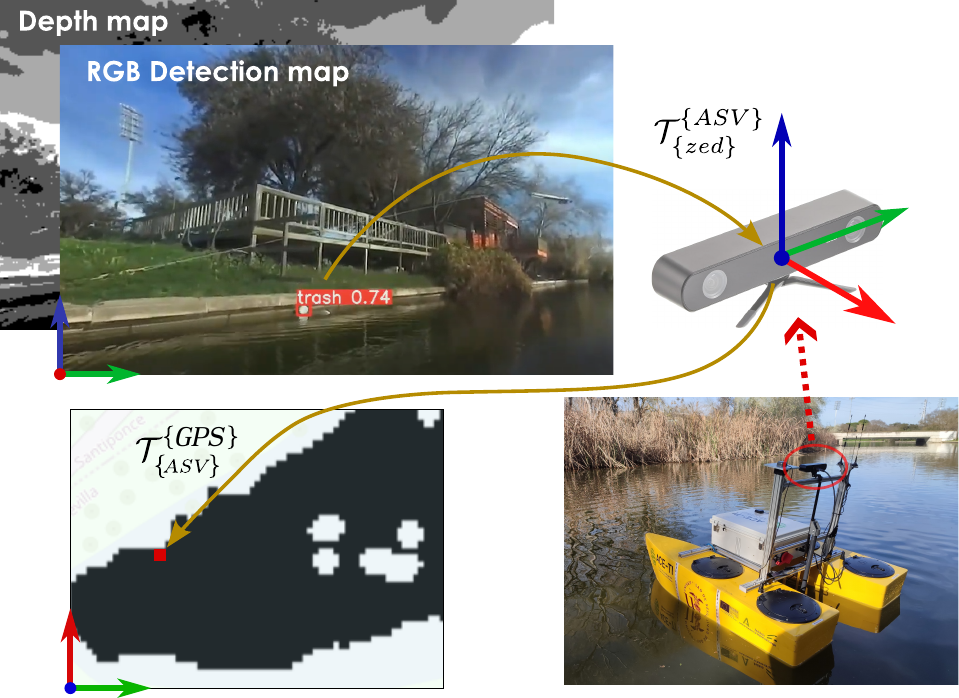}
    \caption{Prototype of a scout ASV equipped with a Zed 2i stereo camera and differential GPS. An example of the YOLOv8 trash detection scheme in a real scenario is presented, showing the bounding box of the detection with its predictive confidence. Depth camera triangulation, GPS and heading enable global waste localization.}
    \label{fig:zed_yolo_scheme}
\end{figure}

The detection of trash for both teams relies on a YOLOv8\footnote{\url{https://github.com/ultralytics/ultralytics}} model, which has been specifically trained on publicly available datasets like \cite{trash_dataset} to recognize plastic debris in aquatic settings with up to 90\% accuracy. For the scout team, this field-tested model processes RGB images captured by the vehicle’s onboard Stereolabs Zed 2i stereo vision depth camera and identifies trash items. Once detected, as seen in Fig. \ref{fig:zed_yolo_scheme}, the position of each trash item is mapped using the triangulation of the stereo camera and the vehicle's differential GPS coordinates and heading, obtaining the global GPS coordinates of the detected waste with a reduced margin of error.
Although cleaning vehicles lack stereo cameras, they are still capable of updating the trash model within their immediate surroundings. As they move through the environment, they use their simpler cameras to detect and confirm trash in close proximity. 

\subsection{Environment and Assumptions}
\label{Environment and Assumptions}
The environment is represented as a grid-based map of size $H\times W$ structured as a connected graph, $G=(V,E)$, where each node $v_{i,j} \in V$ corresponds to a specific location of the grid. Thus, the set of nodes can be defined as \( V = \{ v_{i,j} \mid 1 \leq i \leq H, \, 1 \leq j \leq W \} \). The set of edges that connect adjacent nodes is denoted as \( E \subseteq V \times V \), which represents the possible movements between two adjacent nodes. Here, node adjacency is based on the assumption that the grid is 8 connected, meaning that each node can connect to its surrounding eight neighbors. Therefore, nodes with less than 8 edges indicate the presence of surrounding obstacles. This framework treats time as discrete, meaning that each movement along an edge takes a single time step $t$ to complete. However, scout vehicles operate at twice the speed of cleaners, which means that they move two nodes per time step, while cleaner vehicles move one. To simplify, each mission is set to last a fixed number of steps $T$, within which the objective is to collect as much trash as possible.

The positions of the fleet of $N$ vehicles can be represented as a set $\mathcal{P} = \{ p_n \mid n = 1, 2, \ldots, N \}$, where $p_n$ is the position of the vehicle with the $n$ index. Each of the possible positions a vehicle can take is within a node, such as $p_n = v_{i,j} \in V$. The graph can be represented as a matrix $M[i, j] \in \{0, 1\}$ of size $H\times W$, where $M[i, j] = 1$ if the node $v_{i,j}$ is navigable, and 0 otherwise.

\begin{figure}
    \centering
    \includegraphics[width=0.7\linewidth]{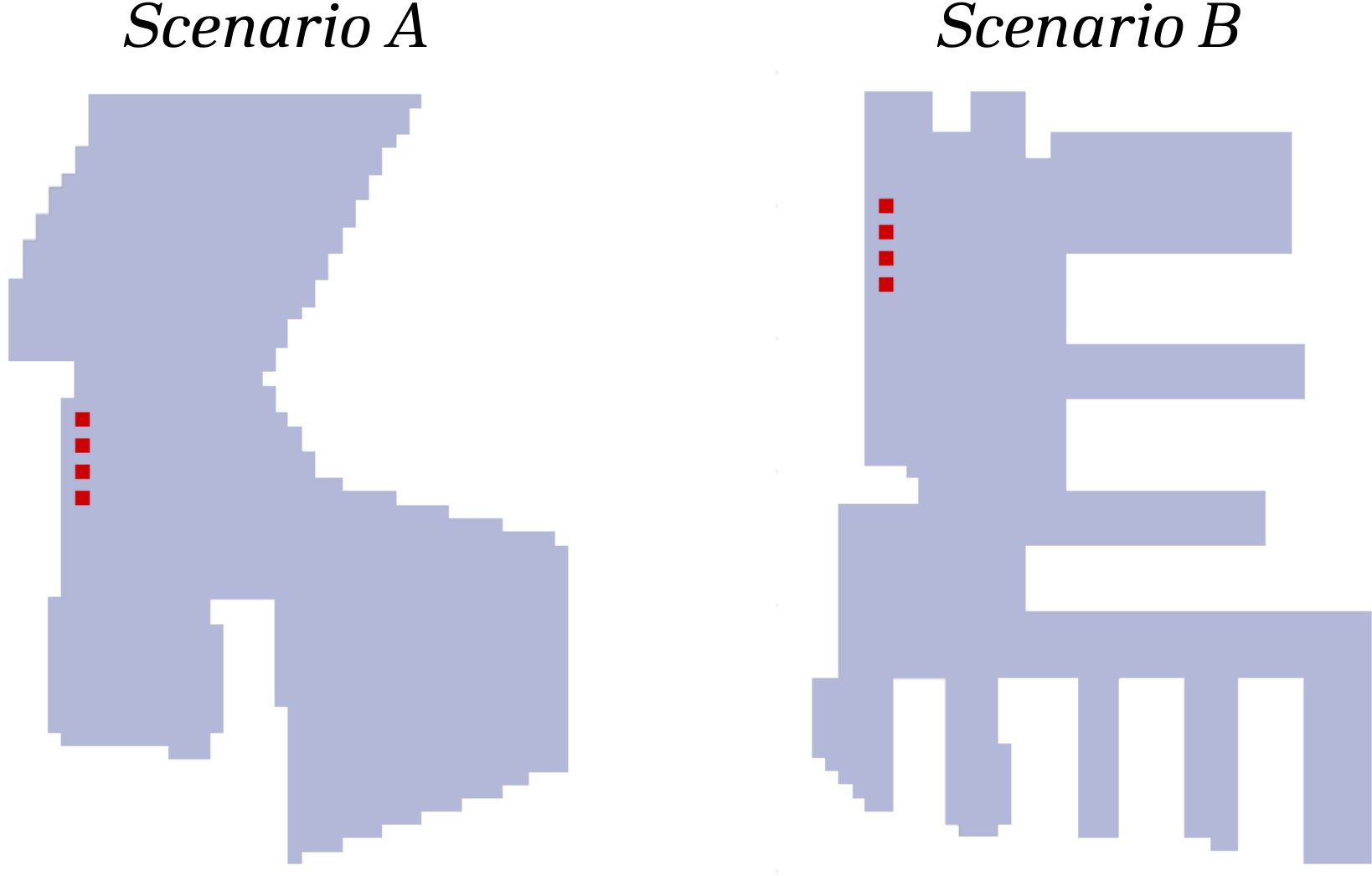}
    \caption{Representation of the two discretized scenario maps which differences in complexity. Initial deployment positions are marked in red.}
    \label{fig:scenarios}
\end{figure}

For simulation purposes, it is necessary to prepare scenarios that allow agents to be trained with characteristics as close as possible to the real ones. For this purpose, two scenario maps with differences in complexity are employed, discretized as matrices $M[i, j]$, something commonly found in similar works such as \cite{barrionuevo2024informative, deng2022automatic}.
The first map is shown on the left in Fig. \ref{fig:scenarios}, and represents a simpler, more open and convex port. In contrast, the second map, shown on the right, depicts a typical sport wharf, with narrow lanes and complex pathways, which introduces greater challenges for the algorithm.
The scenario map and the boundaries are assumed to be known by the agents, and no obstacles will be considered other than those corresponding to boundaries of the map. Thus, possible collisions between vehicles are avoided as they cannot be in the same node.
To ensure a realistic deployment condition, the initial positions of the ASVs will be random within certain safety areas, as seen in  Fig. \ref{fig:scenarios}. Despite this, the algorithm must optimize the coverage of the map with trajectories adapted to these starting points.

The trash positions are defined as a set $B$ with real coordinates in space, following $B = \{ b_k = (x_k, y_k) \mid k = 1, 2, \ldots, K \}$. There, \( K \) is the total number of trash elements generated using a normal distribution at the beginning of the episode, and each element \( b_k \in B \) represents the exact location \( (x_k, y_k) \) where the trash is located in a continuous reference system. 
At the beginning of each episode, a random point from the visitable nodes is selected as the contamination source, around which trash positions are generated using a multivariate normal distribution. Thus, all trash is generated at the start of the episode, and no new items are introduced during the cleaning process. 
In this framework, the items position is dynamically updated at each time step to simulate environmental factors based on two components: wind and random fluctuations. Wind is modeled as a constant velocity $v_{wind}$ for all items, and is defined at the start of the episode sampled from a uniform distribution. Random fluctuations are modeled as random variations in the velocity $v_{rand}$ of each item sampled from uniform distributions at each time step. Then, each trash position is updated following: $b^{t+1}_{k} = b^t_k + \Delta t \left( w_{wind} \cdot \nu_{wind} +  w_{rand} \cdot  \nu_{rand} \right)$. Components $w_{wind}$ and $w_{rand}$ are weights set to 1.

Therefore, the trash distribution map can be defined as a matrix $Y$ of size $H \times W$ where each value $Y[i, j]$ indicates the number of trash positions from $B$ within the corresponding node area at each time step. Thus, $Y[i, j] = \left| \left\{ b_k \in B \mid (x_k, y_k) \in \text{Area}(i, j) \right\} \right|$, where $\text{Area}(i, j)$ represents the area of the cell $(i, j)$ in $M$ matrix. 
Consequently, $Y[i,j] = 0$ implies that the cell [i,j] is free of trash, while a positive value of $Y[i,j]$ indicates the presence of trash items.
When a cleaner agent passes through a cell, all trash items that are placed inside the area of the cell are collected and removed from the $B$ set. It is assumed that the cleaner agents have unlimited carrying capacity, and that the collection process is handled internally by the vehicle's local autopilot and it does not impact the decisions of the proposed algorithm. The amount of trash collected by a cleaner vehicle from $P$ is defined by the number of trash items located in the same cell as the agent at time $t$, i.e., $C(p_n, Y) = Y[i,j] \mid v_{ij} = p_n$.

Due to differences in camera quality between the two teams, the range of coverage they can observe is different. 
It can be defined $\Theta$ as the set of nodes around a vehicle within its field of view, which is determined by a radius $\rho$, following: $\Theta(p_n) = \{ v \in V \mid \norm{v, p_n} < \rho\}$. 
The stereo vision of scout agents enables them to estimate trash positions within a wide range of coverage such that the radius $\rho$ is set to six nodes. The YOLOv8 model processes stereo RGB images captured by the scouts, allowing for high-confidence detection within this radius, as the position of each trash item can be triangulated accurately based on the depth information. In contrast, cleaner agents are equipped with simpler, nonstereo cameras, which restrict their detection capability. Without depth perception, cleaners can only detect trash in immediate proximity. Thus, their vision is limited to a radius $\rho$ of one node around their position, allowing them to detect and clean only nearby trash items. 
This means that the trash model can be defined as a matrix $\hat Y[i,j]$ initialised to 0, where each value is equal to the value of $Y[i,j]$ if an agent $p_n$ is at a distance equal or less than $\rho$. Thus, the model is updated following: $\hat Y[i,j] \leftarrow Y[i,j] \iff v_{ij} \in \Theta(p_n)$.
Both scouts and cleaners are capable of updating this model, although scouts can do so more effectively due to their advanced vision capabilities.

The contamination map is updated at each time step and shared among all agents through a central server, with communication and data processing delays and range being ignored. This share facilitates a cooperative task approach, allowing each agent to benefit from collective knowledge of the environment and improving overall efficiency and coordination.
Based on the range of coverage, a matrix of covered nodes $U[i,j]$ can be defined as a matrix initialised to 0. This matrix is updated at each time step, and each position at a distance equal or less than $\rho$ from the position of a vehicle becomes 1.
 This can be expressed as $U^{t+1}[i, j] \leftarrow \min\left(1, U^t + \Theta(\mathcal{P})\right)$, where each node takes the value of 1 if it has been covered at least once during the episode.

\subsection{Deep Reinforcement Learning}
\subsubsection{State representation}
The framework has been formulated as a Markov Decision Process (MDP), in which each state $s$ represents the current configuration of the agents and the environment. Actions are taken within a set of possible moves $a\in A$, producing a transition from one state to another. For each transition, a reward $r$ will be obtained as a result of the action taken, quantitatively representing how positive that action has been for the overall objective. 
The MDP can be adapted to the multi-agent case by setting it as partially observable, so that the partial state is only accessible by observing agent $n$ through the observation function, such that $o_n = \mathcal{O}(s)$. 
The observation function, $\mathcal{O}(s)$, is a process to map raw input data from a state $s$ to a format that can be interpreted by the agent.

\begin{figure}
\centering
    \includegraphics[width=1\linewidth]{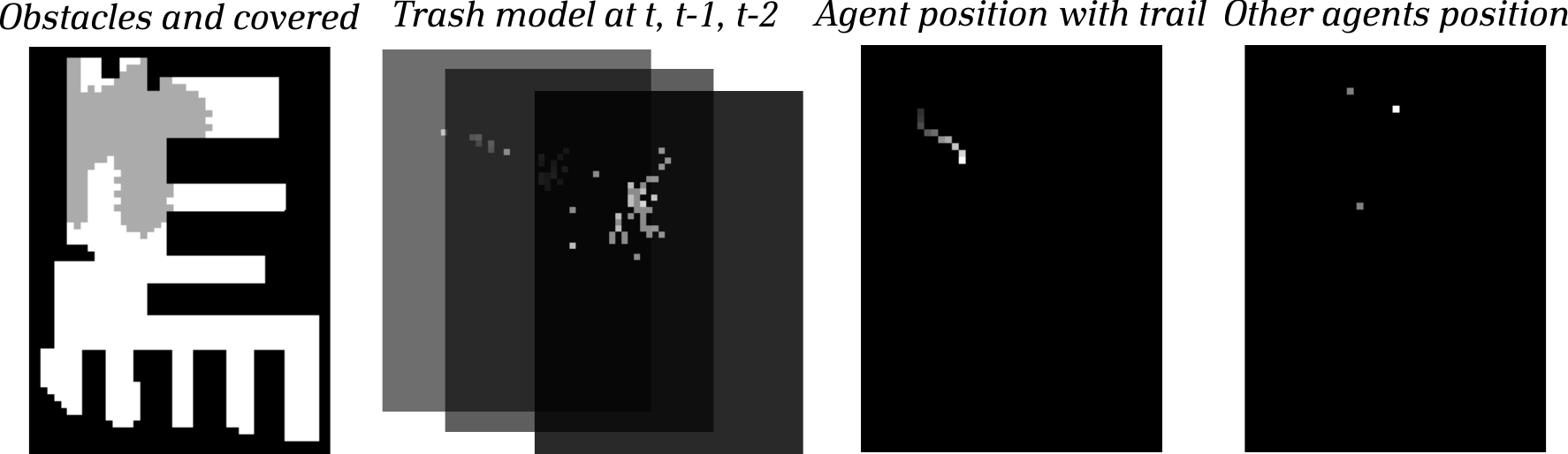}
    \caption{Example of a state representation, composed of six image-like matrices. They are the input to the neural network.}
    \label{fig:states}
\end{figure}

In this approach, the state will be an image-like representation, as it has been proven successful performance in conjunction with convolutional DNNs \cite{samuel2024deepGPS, barrionuevo2024informative, calvo2018heterogeneoustrafficlights}. 
As illustrated in Fig. \ref{fig:states}, this representation is composed of six min-max normalized matrices of size $H \times W$. 
The first corresponds to a matrix with 0 for non-navigable positions, 1 for navigable positions, and 0.5 for navigable positions that have been covered. This matrix can be obtained by $M[i,j]-0.5 \cdot U[i,j]$.
The next three matrices correspond to the trash model matrix $\hat{Y}[i,j]$ at time instants $t$, $t-1$, and $t-2$.
The fifth matrix consists of zeros, except for the ten previous positions visited by the observing agent. This forms a trail with ten values from 0 to 1, creating a fading effect for the recent path of the agent. 
The last matrix consists of zeros, except for the position of other agents. The positions occupied by the scout agents are set to 0.5, and 1 for those with cleaner agents.

\subsubsection{Q-function optimization}
DRL is a type of machine learning in which agents learn to make decisions by interacting with an environment in search of an optimal policy $\pi^*$. Through trial and error, that policy should maximize cumulative reward over time, estimated by the Q-values. In this work Double Deep Q-Learning (DDQL) \cite{DDQL_DSilver} will be employed, a technique used in DRL to improve learning by applying two estimators of the action-value function to mitigate the overestimation of Q-values: Q-network and the Q-target. During the training process, the Q-network updates its parameters iteratively by taking steps to reduce the Bellman error \cite{bellman_eq}, while the Q-target is periodically updated with the weights of the Q-network. Collected experiences will be stored in a prioritized experience replay memory \cite{prioritized_buffer}. Taking samples from this memory, batches are formed to train and update the weights of the network by backpropagation, taking stochastic gradient descent steps toward the direction that reduces the loss $\mathcal{L}$. The loss function is constructed from the quadratic difference between the predicted Q-value from the Q-network and the target Q-value, following:
\begin{equation}\small
\begin{split}
    \mathcal{L}(\vect{\theta}) = \\ \left[R + \gamma \cdot Q^{target}(s', \argmax_{a'}(Q(s';\vect{\theta})); \vect{\theta}^-) - Q(s, a; \vect{\theta})\right]^2
\end{split}
\end{equation}
where $R$ is the immediate reward of the transition, $\gamma$ is the discount factor that weights future rewards, $\vect{\theta}$ are the trainable weights of the Q-network DNN, and $\vect{\theta^-}$ are the frozen parameters of the Q-target DNN. 
In this multi-agent setup, two DNNs are employed, one shared for each team. Their architecture is similar to that used in \cite{samuel2024deepGPS}.
It consists of a first feature extractor of the state representation, formed by three convolutional layers. The output is fed into three fully connected layers, followed by a Dueling Q-Network structure \cite{wang2016dueling}, which decomposes the Q-value estimation into two streams: the value function $V(s)$ and the advantage function $A(s,a)$. This architectural choice forms the basis of the DDDQL algorithm, improving the network's ability to differentiate between the importance of states and actions.

\begin{figure}[t]
\centering
    \includegraphics[width=0.8\linewidth]{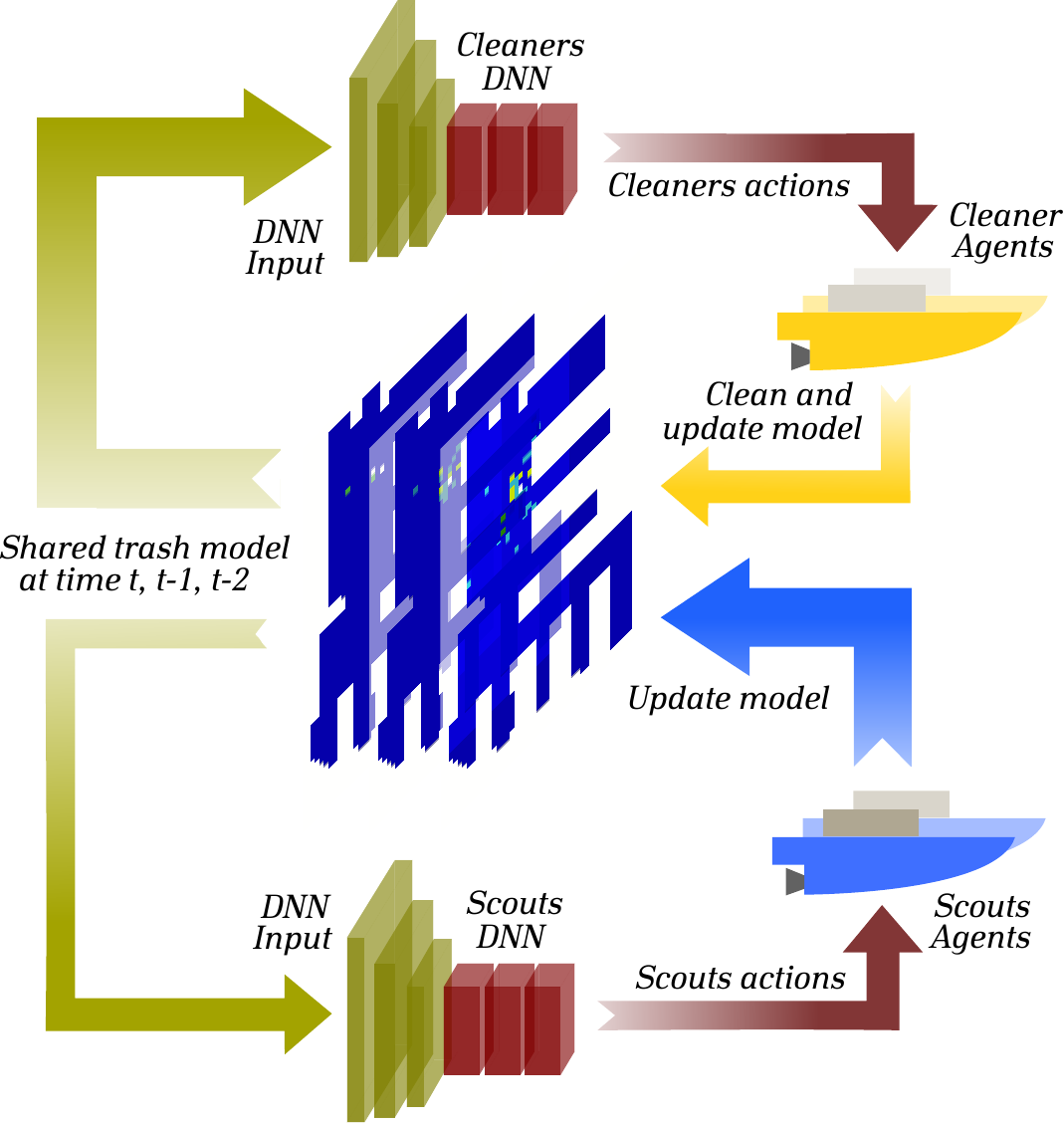}
    \caption{Conceptual diagram of the framework presented in this work. The nexus of cooperation between the two teams is the trash model, which is the input for the DNN of each team. The scout team must provide the updated locations of the waste so that cleaners can collect it. Wider arrows indicate more influence.}
    \label{fig:cooperation_model}
\end{figure}

Reward functions tailored to the specific role of each team have also been designed. Within each team, agent experiences are interchangeable, which enables the share of the prioritized experience replay memory. Additionally, a safety coordination constraint method is employed to prevent collisions, inspired by works like \cite{samuel2024deepGPS}. As seen in Fig. \ref{fig:cooperation_model}, the success of this system is based on cooperation between the two teams. The scout vehicles play a crucial role by providing an accurate and real-time updates on trash locations. In turn, cleaners rely on this information to navigate efficiently to these locations. The better the scouts perform their mapping tasks, the easier it is for the cleaners to complete their work.

\subsubsection{Reward function}\label{Reward function}
The reward function is a crucial element that numerically evaluates the optimality of the actions taken in terms of a final objective. This indicates how beneficial or detrimental a choice is in a specific state. This makes it one of the most important components in DRL, since it is the only feedback available to the agent to assess the value of its actions. Some good reward design principles are explored in \cite{knox2022rewardmisdesignautonomousdriving, sowerby2022designingrewardsfastlearning}. For example, it is stated that penalizing each step taken tends to induce faster learning compared to simply rewarding the achievement of the goal. In addition, rewards should gradually increase as the agent approaches the goal, providing a progressive incentive. 
Following these recommendations, to avoid sparse rewards, the proposed reward functions for each type of agent are designed with these principles in mind, and are defined as follow:
\begin{equation}
\begin{aligned}
    & R_C(s,a_n) = c_{\alpha} \cdot r_{\alpha,n} + c_{\delta} \cdot r_{\delta,n} \\
    & R_S(s,a_n) = c_{\alpha} \cdot r_{\alpha,n} + c_{\beta} \cdot r_{\beta,n} + c_{\gamma} \cdot r_{\gamma,n}\\
\end{aligned}
\end{equation}
where $R_C(s,a_n)$ and $R_S(s,a_n)$ are the reward functions of the cleaner and scout team respectively, and $c_{\alpha}$, $c_{\beta}$, $c_{\gamma}$ and $c_{\delta}$ are parameters to be assessed in order to obtain an effective learning during training. 

Both teams share a common penalization $r_{\alpha, n}$, in terms of the negative distance to the nearest known trash to the agent $n$ if the agent has not cleaned in that step. This can be expressed as: $r_{\alpha,n} = -\min(\norm{p_n, B}) \iff C(n,Y) > 0$. For the cleaner team, each agent is also rewarded for the number of trash items collected, following: $r_{\delta,n} = C(p_n, Y)$.

\begin{figure}
\centering
    \includegraphics[width=0.7\linewidth]{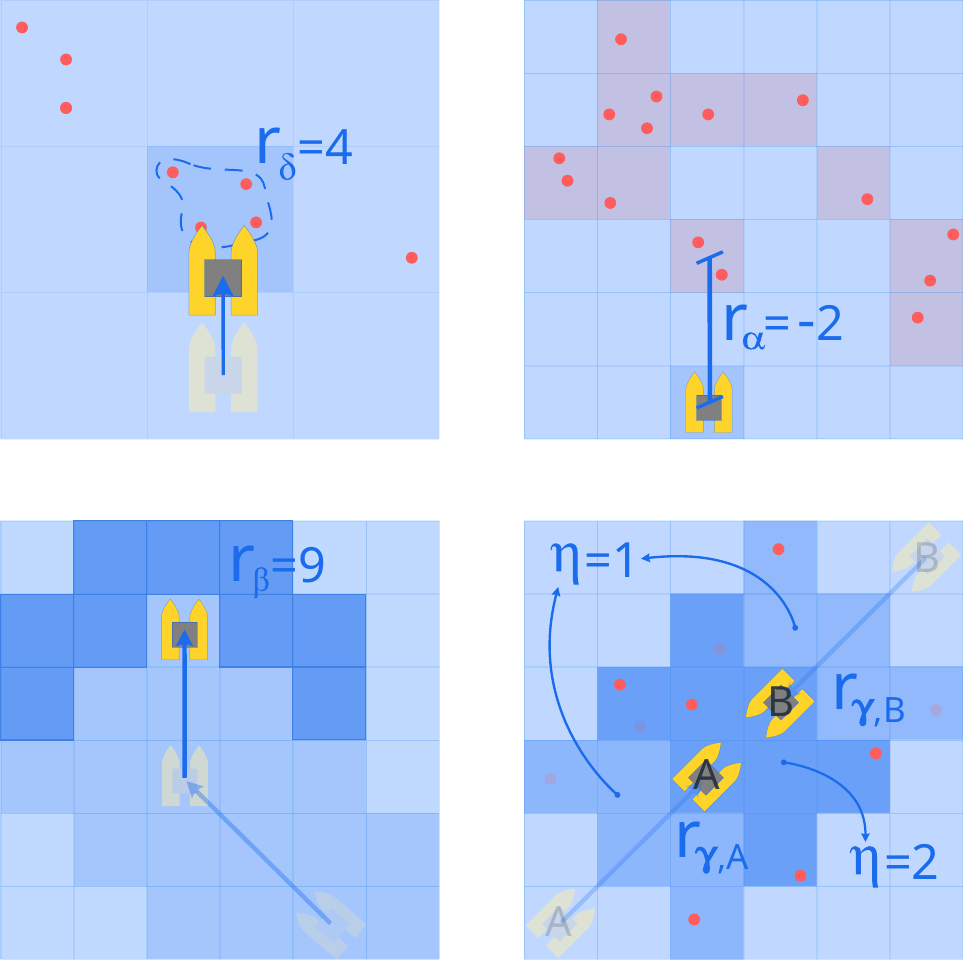}
    \caption{Visual representations of the terms used in the reward functions.}
    \label{fig:rewards}
\end{figure}

For the scout team, there will be two additional rewards to enhance exploration and exploitation. 
To boost an initial exploration phase, they receive a reward $r_{\beta, n}$ for covering non-covered areas, following: $r_{\beta,n} = \sum U^{t} - U^{t-1}$.
This reward will diminish once most of the map has been explored, thus encouraging the rapid identification of high-contamination zones.
The last reward $r_{\gamma,n}$ is defined by the changes produced in the model, calculated as the mismatch between two models within the coverage area. This is expressed as: $r_{\gamma,n} = \sum_{ij} \frac{\hat Y_{ij}^{t} - \hat Y_{ij}^{t-1}}{\eta(v,\mathcal{P})}$, where $\eta(v,\mathcal{P})$ is the number of vehicles $p_n$ simultaneously covering a node $v_{ij}$, defined as: $\eta(v, \mathcal{P}) = \left| \{ p_n \in \mathcal{P} \mid \norm{v, p_n} < \rho \} \right|$. Examples of the proposed reward terms are shown in Fig. \ref{fig:rewards}, where it can be observed how agents move into uncovered areas to obtain $r_{\beta}$ rewards, or that the $\eta$ value changes when the coverage areas of two agents overlap.

\section{Experiments}
\label{Experiments}
In this section, the evaluation of the proposed system is carried out, as well as the definition of the experiments performed. 
Since cleanup conditions can vary significantly, two types of ports have been proposed as evaluation scenarios: Scenario A, characterized by its high convexity and ease of going between points in a straight line, and Scenario B, representing a typical sport wharf design with narrow corridors and challenging access between points, as seen in Fig. \ref{fig:scenarios}. 
Both scenarios are represented by matrices of $H=62 \times W=46$.
For each of them, a comparison will be made between some variations of the proposed approach with four other state-of-the-art benchmark algorithms and heuristics from the literature. Specifically, the algorithms employed correspond to lawn mower, random walker, PSO and greedy. To assess cooperation within and between teams, training and simulations will be conducted with two agents per team. In this approach, this number could be scaled up with minimal impact on computational cost, as agents within the same team share experiences and use a common neural network.
All simulations and trainings have been conducted on a workstation running Ubuntu 22.04, equipped with a GPU Nvidia RTX 4090 25GB. PyTorch libraries are used to define DNNs, and the code is available on Github \url{https://github.com/amendb/HeterogeneousTrashCleanup} for reproducing the results.

Due to the off-policy nature of DDDQL, this algorithm may benefit from using simpler algorithms to learn from certainly successful actions, instead of basing learning entirely on random decisions. For this reason, two types of training will be tested: one using $\epsilon$-greedy  policies, where actions are selected entirely at random, and another in which half of the actions are chosen randomly, while the other half are selected by a greedy algorithm. The greedy algorithm, which has demonstrated notably good performance, may offer a significant advantage by helping the agent learn more effectively from successful actions, but random actions are still employed as it is also important for the agent to continue observing negative actions in order to understand which choices lead to bad rewards. Additionally, in the latter case, 20\% of the buffer will be pre-filled at the start of the training with experiences collected from agents using the greedy algorithm. This percentage has been evaluated through trial and error, and is expected to help improve the initial learning process and guide the agent towards more effective policies from the start.

The reward function ponderations defined in Section \ref{Reward function} were determined by trial and error until satisfactory results were achieved, establishing $c_{\alpha} = 1$, $c_{\delta} = 50$, $c_{\beta}= 2$, $c_{\gamma} = 1$.
While hyperparameter tuning might yield slightly better results, this is outside of the scope of this article. With fixed hyperparameters DRL can be applied across a wide range of scenarios without the need for prior knowledge. Its strength lies in its model-free nature, which enables it to generalize effectively across diverse situations, such as waste movement. 

As mentioned in Section \ref{Environment and Assumptions}, in each episode the four agents appear at four initial positions on the map, with a sufficient level of autonomy to complete a mission of 150 movements, regardless of the distance traveled. The number of trash items is generated using a normal distribution $\mathcal{N}(60,10^2)$. Each training process has a duration of $60,000$ episodes, in which the number of garbage elements, their positions, and the flows that move them will be generated randomly.
Transitions from each team will be stored in a prioritized experience replay memory capable of storing up to $1 \times10^{6}$ experiences. From memory batches of size 128 will be sampled, to adjust the network weights with a learning rate of $1\times10^{-4}$. 
The policy chosen is the one that obtains the best performance on average.
For the evaluation process, the average results of the same 100 episodes have been recorded for each algorithm. This diversity is what allows for generalization in the evaluation of the effectiveness of the algorithms. 
These episodes have never been seen during the DRL training process.
Two metrics will be employed to quantitatively measure the performance of them:
\begin{itemize}
    \item \textbf{PTC}: The percentage of trash cleaned (PTC) over time, that quantifies the effectiveness of the cleaning algorithm. It is calculated as the ratio of the final amount of trash removed by the subset of cleaners agents from $\mathcal{P}$ to the total amount of trash present in the environment $K$ at the beginning of the episode.
    \[PTC(t) = \frac{\sum_{\tau=0}^t C(\mathcal{P},Y^{\tau})}{K}\cdot 100\]

    \item \textbf{MSE}: Mean Squared Error over time between the model $\hat Y$ and the trash ground truth $Y$. As $Y,\hat Y$ are sparse matrices and the elementwise error will produce high error variations, a Gaussian Filter $G$ with $\sigma = 1$ is applied to compare the density of the trash distribution instead:
    \[MSE(t) = \frac{1}{H \times W} \sum_{\tau=0}^t\left(G \circledast Y^{\tau} - G \circledast  \hat Y^{\tau}\right)^2\]

\end{itemize}

\begin{figure}[t]
\centering
    \includegraphics[width=1\linewidth]{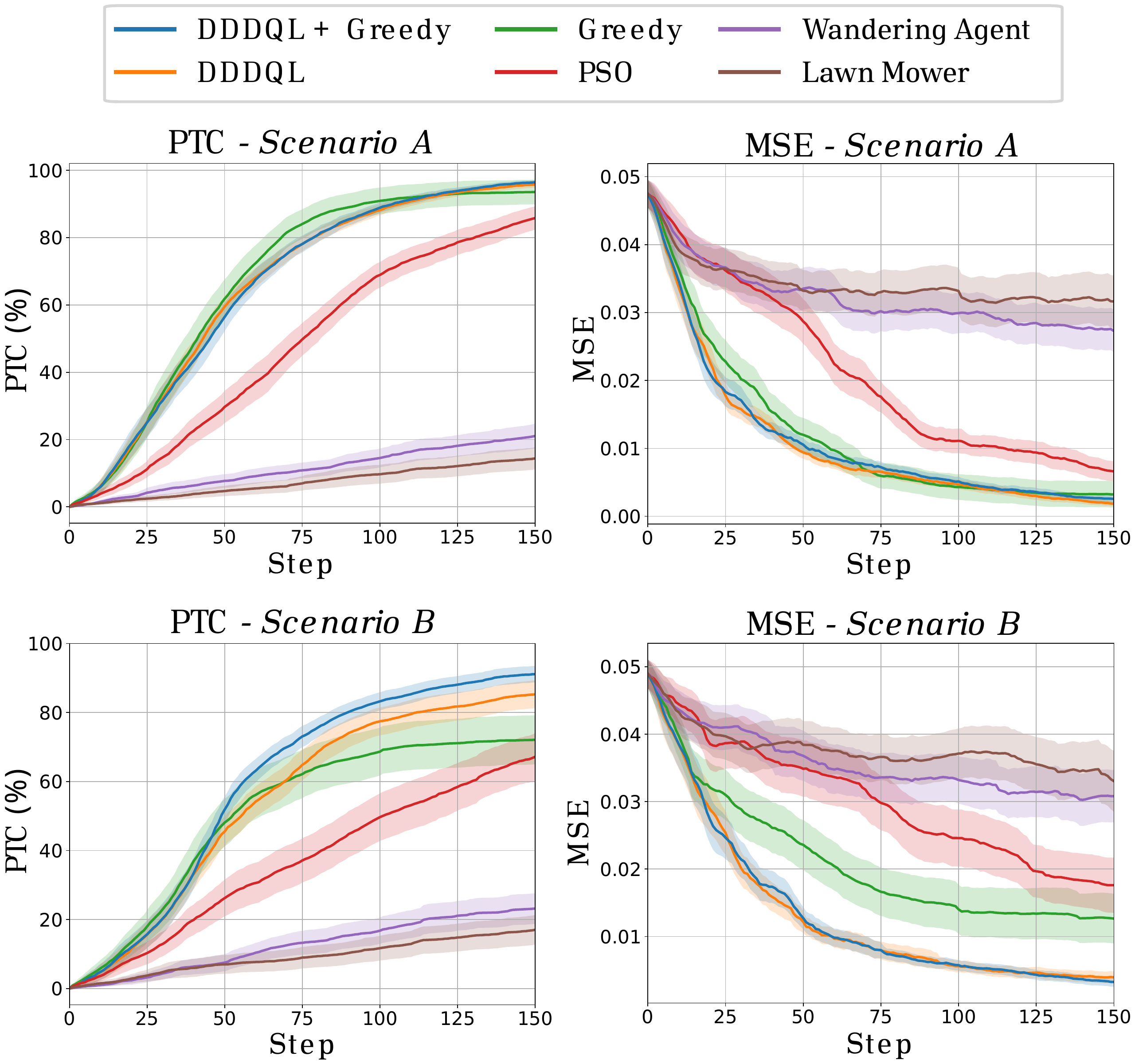}
    \caption{Graphical comparison over time of the percentage of trash cleaned (left) and mean squared error (right) for each of the benchmark algorithms, with confidence intervals represented as shaded regions. The upper graphs correspond to Scenario A and the lower graphs correspond to Scenario B.}
    \label{fig:results}
\end{figure}

\begin{table*}[thb]
\scriptsize
\caption{Metrics comparison between DDDQL trainings and other algorithms at the end of a evaluation set of 100 episodes. Highlighted values correspond to the best performance value. Average computation time is also included.}
\label{tab:algorithms_comparative}
\centering
\setlength{\tabcolsep}{5pt} 
\renewcommand{\arraystretch}{1.5}
\begin{tabular}{|c||cc|cc||cc|cc||c|}
\hline
  \multirow{3}{*}{\textbf{Algorithm}} & 
  \multicolumn{4}{c||}{\textbf{Scenario A}} &
  \multicolumn{4}{c||}{\textbf{Scenario B}} & 
   \multirow{3}{*}{\makecell{\textbf{Average time of}\\ \textbf{computation ($ms$)}}} \\ 
  \cline{2-9} 
  & \multicolumn{2}{c|}{\textbf{$\textbf{\text{PTC}}$}} &
  \multicolumn{2}{c||}{\textbf{$\textbf{\text{MSE}}$}}  & 
  \multicolumn{2}{c|}{\textbf{$\textbf{\text{PTC}}$}} &
  \multicolumn{2}{c||}{\textbf{$\textbf{\text{MSE}}$}} & \\ 
  & Mean       & CI $95\%$       & Mean       & CI $95\%$   & Mean       & CI $95\%$       & Mean       & CI $95\%$ & \\ \hline\hline
DDDQL & 95.83 & 0.94 & \textbf{0.0018} & 0.0004 & 85.28 & 4.00 & 0.0040 & 0.0009 & 19.24
   \\ \hline\hline
DDDQL + Greedy & \textbf{96.49} & 0.67 & 0.0026 & 0.0005 & \textbf{91.13} & 2.38 & \textbf{0.0032} & 0.0007 & 19.36
   \\ \hline\hline
Random Walker & 21.07 & 3.67 & 0.0272 & 0.0031 & 23.19 & 4.42 & 0.0308 & 0.0039 & 0.15
   \\ \hline\hline
Lawn Mower & 14.41 & 3.28 & 0.0316 & 0.0038 & 17.03 & 4.29 & 0.0330 & 0.0045 & 0.15
   \\ \hline\hline
PSO & 85.82 & 3.45 & 0.0066 & 0.0015 & 67.19 & 6.80 & 0.0176 & 0.0041 & 0.36
   \\ \hline\hline
Greedy & 93.56 & 3.66 & 0.0032 & 0.0019 & 72.11 & 7.10 & 0.0126 & 0.0037 & 1.33
   \\ \hline
\end{tabular}

\end{table*}

The evaluation results can be observed in Fig. \ref{fig:results}, where the performance of the algorithms is visualized across key metrics along the steps, and in Table \ref{tab:algorithms_comparative}, where the final average final metrics are summarized and compared.
The results show, as expected, that the algorithms that do not take into account the mission objective and move following a fixed pattern (lawn mower and random walker) perform significantly worse than the other approaches, both in terms of average MSE and PCT. 
The PSO algorithm ranks third overall, sometimes nearly matching the performance of the Greedy algorithm, particularly in Scenario B.

The three top-performing algorithms are clearly the two DRL proposals and the Greedy algorithm, the latter being especially remarkable for its good performance despite its simplicity. It stands out for its speed in achieving high cleanliness percentages soon, making it a competitive option in terms of efficiency and challenging DRL algorithms in Scenario A, where the open layout allows simpler algorithms to achieve remarkable results. However, supported by the metrics shown in Table \ref{tab:algorithms_comparative}, while the Greedy algorithm stagnates at that early high PTC, DRL-based algorithms consistently demonstrate better final performance across both scenarios. In particular, in Scenario B, characterized by its complex structure and intricate pathways, the best DRL approach outperforms the Greedy algorithm by up to 26\% in average PTC, demonstrating superior adaptability in more demanding tasks\footnote{See the attached video to dynamically visualize these explained behaviors.}.

In terms of MSE, it is important to keep in mind that as more trash is cleaned up, the MSE tends to decrease naturally, since the model to be estimated becomes simpler. The DRL-based algorithms consistently deliver clearly superior results compared to other methods, as seen in Table \ref{tab:algorithms_comparative}. The values achieved by the DRL approaches are significantly lower, highlighting their ability to generate more accurate representations of the environment. This trend is further reinforced by the smaller confidence intervals (CI) observed in the DRL results, indicating that the DRL algorithms not only perform better on average but also exhibit less variability between different episodes, making them more reliable across diverse scenarios.

\begin{figure}
    \centering
    \includegraphics[width=0.9\linewidth]{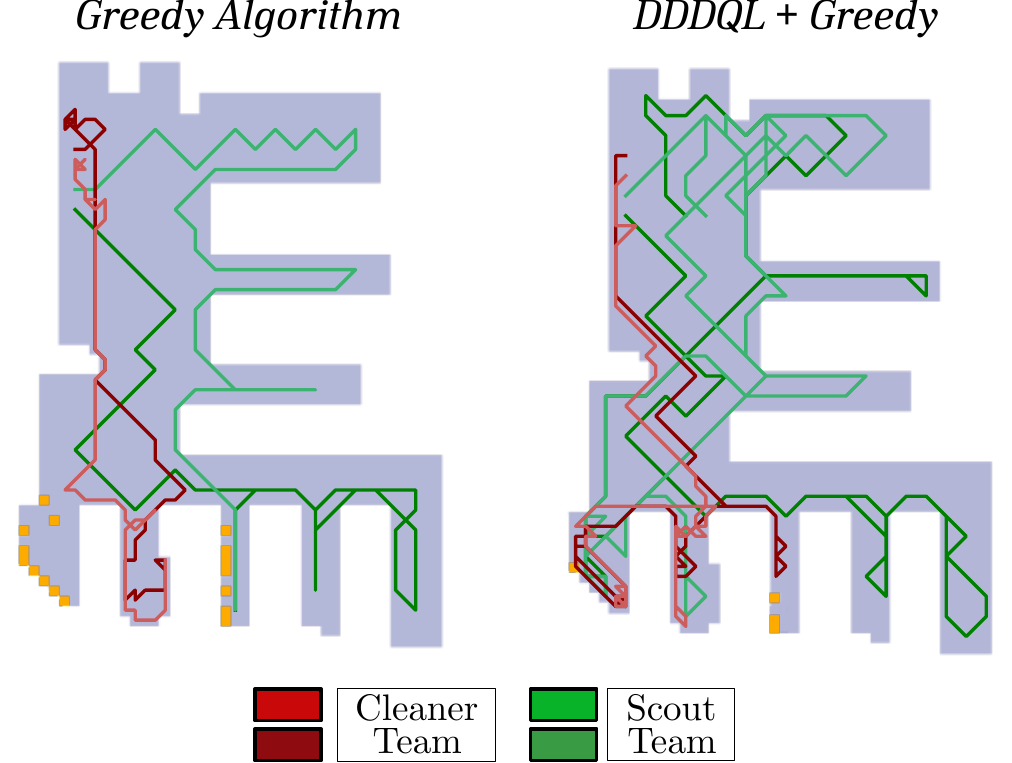}
    \caption{Comparison paths in a random episode at Scenario B between Greedy and DDDQL+Greedy algorithms. Nodes with trash remaining at the end of the episode are marked in yellow.}  \label{fig:path_comparison}
\end{figure}

In contrast, the Greedy algorithm, while highly efficient in targeting waste and achieving high cleanliness percentages, is limited by its simplistic nature. The Greedy approach follows a very direct path to known waste locations, which makes it fast but prevents it from understanding or navigating the more complex boundaries and obstacles within the map. This limitation is especially evident in Scenario B, where the Greedy algorithm struggles due to the presence of narrow corridors and difficult-to-reach areas. In comparison, DRL algorithms excel at learning optimized trajectories, allowing them to adapt to complex environments and efficiently handle boundaries, even when waste is distributed in hard-to-reach locations. This can be observed in Fig. \ref{fig:path_comparison}, where the DRL approach manages to scan most of the port's corridors, and the cleaners collect almost all of the trash although they are located in areas difficult to access, managing to make trajectories that skirt the borders. In contrast, it is observed how the cleaners of the Greedy algorithm get stuck in a corridor and are not able to exit to another corridor where the scouts have discovered waste.
However, one challenge that DRL-based methods may encounter is precision; in some cases, they may not target specific waste nodes as directly as the Greedy algorithm. Greedy algorithm is known to provide near-optimal solutions with good approximation ratios, especially under matroid constraints in submodular functions \cite{GreedIsGood}, making it ideal for tasks requiring rapid convergence. However, the scalability and adaptability of DRL algorithms provide a clear advantage in more dynamic and obstacle-laden environments.

Focusing exclusively on the comparison between the two DRL algorithms, in terms of PTC, DDDQL trained with Greedy actions consistently outperforms the standalone DDDQL across both scenarios, with this difference being more pronounced in Scenario B. However, when looking at the MSE metric, the results are more comparable between the two approaches, and in some cases, the standalone DDDQL performs slightly better, as observed in Scenario A. This seems to indicate that when the task is more complex training with Greedy actions tends to result in better performance, such as cleaning (PTC metric) is harder than explore (MSE metric), or navigating through the more intricate layout of Scenario B. In contrast, when the task is simpler with less restrictive boundaries, as in Scenario A, the trial-and-error approach with random actions during the DDDQL algorithm's training proves sufficient to achieve similar results without the need for Greedy actions.

Table \ref{tab:algorithms_comparative} also includes the inference times for each algorithm to choose an action for the entire fleet, measured in milliseconds. It can be observed that the DRL-based approaches have significantly higher computational times compared to the other methods, probably due to the complex decision-making process involved in DRL using DNNs. However, the higher computational cost of DRL methods is compensated by their higher adaptability and performance in complex scenarios. Furthermore, assuming an average ASV speed of between 1 and 2 meters per second, depending on the type of agent, inference times in the order of milliseconds can be negligible in a real implementation.

\section{Conclusion}
\label{Conclusion}
This paper presents a model-free DRL-based framework for IPP with heterogeneous fleets of ASVs for floating waste collection. The proposed system utilizes two teams of ASVs: scouts, responsible for mapping the trash distribution, and cleaners, in charge of waste collection. Coordination between these teams is achieved through DRL algorithms, allowing the agents to learn optimal strategies to maximize cleaning efficiency. Experimental results demonstrate the effectiveness of the proposed approach, outperforming benchmark algorithms in two distinct scenarios representing different cleanup conditions. The DRL-based algorithms exhibit specially superior adaptability in complex environments, achieving higher cleaning percentages and generating more accurate representations of the trash distribution. Additionally, training with Greedy actions further enhances the performance of DRL, particularly in scenarios with intricate layouts. 

However, in less complex scenarios the differences between the DRL-based methods and the Greedy algorithm are not so significant. Greedy algorithms require a prior study of the environment to determine their applicability and need to be adjusted for each specific situation. On the contrary, DRL does not require prior knowledge of the environment or adaptation of heuristics, due to its model-free nature, and allows to generalize effectively in a variety of situations.

In addition, while DRL-based methods have higher computational times compared to simpler approaches, their enhanced adaptability and performance capabilities compensate for this computational cost. 
Furthermore, in real-world applications with ASVs running at typical speeds, the millisecond inference times associated with DRL computation are negligible.

\footnotesize
\bibliographystyle{ieeetr} 
\bibliography{main} %

\begin{thebibliography}{10}

\bibitem{plasticfish2016}
{World Economic Forum, Ellen MacArthur Foundation and McKinsey \& Company}, ``{The New Plastics Economy: Rethinking the future of plastics},'' 2016.

\bibitem{advancesinmultirobotsurvey}
Y.~Cai and S.~X. Yang, ``A survey on multi-robot systems,'' in {\em World Automation Congress 2012}, pp.~1--6, IEEE, 2012.

\bibitem{rizk2019cooperative}
Y.~Rizk, M.~Awad, and E.~W. Tunstel, ``Cooperative heterogeneous multi-robot systems: A survey,'' {\em ACM Computing Surveys (CSUR)}, vol.~52, no.~2, pp.~1--31, 2019.

\bibitem{hernandez2019survey}
P.~Hernandez-Leal, B.~Kartal, and M.~E. Taylor, ``A survey and critique of multiagent deep reinforcement learning,'' {\em Autonomous Agents and Multi-Agent Systems}, vol.~33, no.~6, pp.~750--797, 2019.

\bibitem{popovic2024learning}
M.~Popović, J.~Ott, J.~Rückin, and M.~J. Kochenderfer, ``Learning-based methods for adaptive informative path planning,'' {\em Robotics and Autonomous Systems}, vol.~179, p.~104727, 2024.

\bibitem{micaela2022aquafelpso}
M.~{Jara Ten Kathen}, F.~Peralta, P.~Johnson, I.~{Jurado Flores}, and D.~G. Reina, ``Aquafel-pso: An informative path planning for water resources monitoring using autonomous surface vehicles based on multi-modal pso and federated learning,'' {\em Ocean Engineering}, vol.~311, p.~118787, 2024.

\bibitem{Federico57.2023}
F.~Peralta, D.~G. Reina, and S.~L. Toral, ``Water quality online modeling using multi-objective and multi-agent bayesian optimization with region partitioning,'' {\em Mechatronics}, vol.~91, p.~102953, 2023.

\bibitem{drlpathplanning}
M.~Grzelczak and P.~Duch, ``Deep reinforcement learning algorithms for path planning domain in grid-like environment,'' {\em Applied Sciences}, vol.~11, no.~23, p.~11335, 2021.

\bibitem{vashisth2024deep}
A.~Vashisth, J.~Rückin, F.~Magistri, C.~Stachniss, and M.~Popović, ``Deep reinforcement learning with dynamic graphs for adaptive informative path planning,'' {\em IEEE Robotics and Automation Letters}, vol.~9, no.~9, pp.~7747--7754, 2024.

\bibitem{lei2018dynamic}
X.~Lei, Z.~Zhang, and P.~Dong, ``Dynamic path planning of unknown environment based on deep reinforcement learning,'' {\em Journal of Robotics}, vol.~2018, no.~1, p.~5781591, 2018.

\bibitem{samuel2024deepGPS}
S.~Yanes~Luis, D.~Shutin, J.~Marchal~Gómez, D.~G. Reina, and S.~L. Toral, ``Deep reinforcement multiagent learning framework for information gathering with local gaussian processes for water monitoring,'' {\em Advanced Intelligent Systems}, vol.~6, no.~8, p.~2300850, 2024.

\bibitem{yang2020multi}
Y.~Yang, L.~Juntao, and P.~Lingling, ``Multi-robot path planning based on a deep reinforcement learning dqn algorithm,'' {\em CAAI Transactions on Intelligence Technology}, vol.~5, no.~3, pp.~177--183, 2020.

\bibitem{woo2020collision}
J.~Woo and N.~Kim, ``Collision avoidance for an unmanned surface vehicle using deep reinforcement learning,'' {\em Ocean Engineering}, vol.~199, p.~107001, 2020.

\bibitem{gu2017deep}
S.~Gu, E.~Holly, T.~Lillicrap, and S.~Levine, ``Deep reinforcement learning for robotic manipulation with asynchronous off-policy updates,'' in {\em 2017 IEEE international conference on robotics and automation (ICRA)}, pp.~3389--3396, IEEE, 2017.

\bibitem{barrionuevo2024informative}
A.~M. Barrionuevo, S.~Yanes~Luis, D.~G. Reina, and S.~L. Toral, ``Informative deep reinforcement path planning for heterogeneous autonomous surface vehicles in large water resources,'' {\em IEEE Access}, vol.~12, pp.~71835--71852, 2024.

\bibitem{calvo2018heterogeneoustrafficlights}
J.~A. Calvo and I.~Dusparic, ``Heterogeneous multi-agent deep reinforcement learning for traffic lights control,'' in {\em Irish Conference on Artificial Intelligence and Cognitive Science}, vol.~2259 of {\em {CEUR} Workshop Proceedings}, pp.~2--13, CEUR-WS.org, 2018.

\bibitem{gao2023asymmetric}
Y.~Gao, J.~Chen, X.~Chen, C.~Wang, J.~Hu, F.~Deng, and T.~L. Lam, ``Asymmetric self-play-enabled intelligent heterogeneous multirobot catching system using deep multiagent reinforcement learning,'' {\em IEEE Transactions on Robotics}, vol.~39, no.~4, pp.~2603--2622, 2023.

\bibitem{akib2019unmanned}
A.~Akib, F.~Tasnim, D.~Biswas, M.~B. Hashem, K.~Rahman, A.~Bhattacharjee, and S.~A. Fattah, ``Unmanned floating waste collecting robot,'' in {\em TENCON 2019 - 2019 IEEE Region 10 Conference (TENCON)}, pp.~2645--2650, 2019.

\bibitem{kamarudin2021development}
N.~Kamarudin, I.~N.~A. Mohd~Nordin, D.~Misman, N.~Khamis, M.~Razif, and F.~Hanim, ``Development of water surface mobile garbage collector robot,'' {\em Alinteri Journal of Agriculture Sciences}, vol.~36, pp.~534--540, 06 2021.

\bibitem{deng2022automatic}
T.~Deng, X.~Xu, Z.~Ding, X.~Xiao, M.~Zhu, and K.~Peng, ``{Automatic collaborative water surface coverage and cleaning strategy of UAV and USVs},'' {\em Digital Communications and Networks}, 12 2022.
\newblock doi: 10.1016/j.dcan.2022.12.014.

\bibitem{trash_dataset}
Y.~Cheng, J.~Zhu, M.~Jiang, J.~Fu, C.~Pang, P.~Wang, K.~Sankaran, O.~Onabola, Y.~Liu, D.~Liu, and Y.~Bengio, ``Flow: A dataset and benchmark for floating waste detection in inland waters,'' in {\em 2021 IEEE/CVF International Conference on Computer Vision (ICCV)}, pp.~10933--10942, 2021.

\bibitem{DDQL_DSilver}
H.~Van~Hasselt, A.~Guez, and D.~Silver, ``Deep reinforcement learning with double q-learning,'' in {\em Proceedings of the AAAI conference on artificial intelligence}, vol.~30, p.~2094–2100, 2016.

\bibitem{bellman_eq}
R.~Bellman, {\em Dynamic Programming}.
\newblock Princeton University, NJ, USA: Princeton University Press, 1~ed., 1957.

\bibitem{prioritized_buffer}
T.~Schaul, J.~Quan, I.~Antonoglou, and D.~Silver, ``Prioritized experience replay,'' in {\em 4th International Conference on Learning Representations (ICLR)}, 2016.

\bibitem{wang2016dueling}
Z.~Wang, T.~Schaul, M.~Hessel, H.~van Hasselt, M.~Lanctot, and N.~de~Freitas, ``Dueling network architectures for deep reinforcement learning,'' in {\em 33rd International Conference on Machine Learning (ICML)}, pp.~1995--2003, 2016.

\bibitem{knox2022rewardmisdesignautonomousdriving}
W.~B. Knox, A.~Allievi, H.~Banzhaf, F.~Schmitt, and P.~Stone, ``Reward (mis)design for autonomous driving,'' {\em Artificial Intelligence}, vol.~316, p.~103829, 2023.

\bibitem{sowerby2022designingrewardsfastlearning}
H.~Sowerby, Z.~Zhou, and M.~L. Littman, ``Designing rewards for fast learning.''
\newblock arXiv preprint arXiv:2205.15400, 2022. Available: https://arxiv.org/abs/2205.15400.

\bibitem{GreedIsGood}
M.~Feldman, C.~Harshaw, and A.~Karbasi, ``Greed is good: Near-optimal submodular maximization via greedy optimization,'' in {\em Proceedings of the 2017 Conference on Learning Theory}, vol.~65, pp.~758--784, 2017.

\end{thebibliography}

\end{document}